%% file: ethics.tex
\pdfoutput=1

\documentclass[11pt]{article}

\usepackage[]{naacl2021}

\usepackage{times}
\usepackage{latexsym}

\usepackage[T1]{fontenc}

\usepackage[utf8]{inputenc}
\usepackage{enumerate}
\usepackage{microtype}
\usepackage{csquotes}
\usepackage{multirow}
\usepackage{booktabs}
\usepackage{todonotes}

\newcommand{\cfr}{\textit{Final Rule}}

\title{Beyond Fair Pay: \\Ethical Implications of NLP Crowdsourcing}

\author{Boaz Shmueli$^{1,2,3,}$\thanks{~~Corresponding author: \href{mailto:shmueli@iis.sinica.edu.tw?Subject=Reactive Supervision: A New Method for Collecting Sarcasm Data}{shmueli@iis.sinica.edu.tw}}~,~ Jan Fell$^2$,~ Soumya Ray$^2$, \and Lun-Wei Ku$^3$\\
$^1$Social Networks and Human-Centered Computing, TIGP, Academia Sinica\\
$^2$Institute of Service Science, National Tsing Hua University\\
$^3$Institute of Information Science, Academia Sinica\\
}
\begin{document}
\maketitle
\begin{abstract}
The use of crowdworkers in NLP research is growing rapidly, in tandem with the exponential increase in research production in machine learning and AI. Ethical discussion regarding the use of crowdworkers within the NLP research community is typically confined in scope to issues related to labor conditions such as fair pay. We draw attention to the lack of ethical considerations related to the various tasks performed by workers, including labeling, evaluation, and  production. We find that the \cfr{}, the common ethical framework used by researchers, did not anticipate the use of online crowdsourcing platforms for data collection, resulting in  gaps between the spirit and practice of human-subjects ethics in NLP research. We enumerate common scenarios where crowdworkers performing NLP tasks are at risk of harm. We thus recommend that researchers evaluate these risks by considering the  three ethical principles set up by the Belmont Report. We also clarify some common misconceptions regarding the Institutional Review Board (IRB) application. We hope this paper will serve to reopen the  discussion within our community regarding the ethical use of crowdworkers.
\end{abstract}
\section{Revisiting the Ethics of Crowdsourcing}
\label{sec:revisiting}
The information age brought with it the internet, big data, smartphones, AI, and along with these, a plethora of complex ethical challenges.
As a result, there is growing concern and discussion on ethics within the research community at large, including the NLP community.
This is manifested in new ethics-focused workshops, ethics conference panels and relevant updates to peer review forms.

While ethics in NLP has multiple aspects, most recent attention focuses on  pressing issues related to
 the societal impact of NLP. These include discrimination, exclusion, over-generalization, bias, and fairness \cite{hovy-spruit-2016-social,leidner-plachouras-2017-ethical}.
Other works are concerned with the ethical implications of NLP shared tasks \cite{parra-escartin-etal-2017-ethical}, 
and introducing ethics into the NLP curriculum \cite{bender-etal-2020-integrating}.

A substantial amount of NLP research now takes advantage of \textit{crowdworkers} --- workers on  crowdsourcing platforms such as 
Amazon Mechanical Turk (known also as AMT or MTurk), Figure Eight\footnote{Previously CrowdFlower; acquired by Appen in 2019.}, Appen, Upwork, Prolific, Hybrid, Tencent Questionnaire, and Baidu Zhongbao, as well as internal crowdsourcing platforms in companies such as Microsoft
and Apple. Workers are recruited to label, evaluate, and produce data. In the pre-internet era, such tasks (e.g. part-of-speech (POS) tagging) were done by hiring expert annotators or linguistics students. However,
these are now mostly replaced by crowdworkers due to lower costs, convenience, speed, and scalability.
\input{table1}

Overall, the general consensus in the literature is that as long as the pay to the crowdworkers is ``fair'' (minimum hourly wage or above), there are no further ethical concerns, and there is no need for approval by an
Institutional Review Board\footnote{Institutional Review Boards (IRBs) are university-level, multi-stakeholder committees that
review the methods proposed for research involving human subjects to ensure that they conform to ethical principles.
IRBs are also known by various other names, such as Research Ethics Boards (REBs) and Research Ethics Committees (RECs). Non-academic organizations may employ  similar committees.}  (with some exceptions).
For example, \citet{hovy-spruit-2016-social} mention that ``[w]ork on existing corpora is unlikely to raise
any flags that would require an IRB approval'', with a footnote that there are ``a few exceptions''. 
\citet{fort2011amazon} mention that  only ``[a] small number of universities have insisted on institutional review board approval for MTurk experiments''. 
As another example, NLP students are being taught that ``paid labeling does not require IRB approval'' since ``[i]t's not an experiment with human subjects'' \cite{cmu-course}. Indeed, NLP papers that
involve crowdsourced work rarely mention a review by an ethics board.

In this work, we wish to revisit the ethical issues of crowdsourcing in the NLP context, highlight several issues of concern, and suggest ways forward. From our survey of top NLP conferences, we find that crowdsourcing tasks are growing rapidly within the community. We therefore establish a common understanding of research ethics and how it relates to crowdsourcing. We demonstrate that the existing ethical framework is often inadequate and does not seek to protect crowdsourced workers. We then dispel common misunderstandings regarding the IRB process that NLP researchers might harbor. And finally, we outline how to apply ethical values as guidelines to minimize potential harms, and conclude with recommendations.

\section{The Rise and Rise of NLP Crowdsourcing}
\label{sec:growing}
To get a sense of the extent and growth of crowdsourced tasks within the NLP research community, we analyzed the proceedings of three top NLP conferences: ACL, EMNLP, and NAACL (also known as NAACL-HLT)\footnote{ACL is the Annual Meeting of the Association for Computational Linguistics, EMNLP is the Conference on Empirical Methods in Natural Language Processing, and NAACL is the Annual Conference of the North American Chapter of the Association for Computational Linguistics: Human Language Technologies.}. We scanned the annual proceedings of these conferences in the six years from 2015 to 2020, looking for papers that mention direct employment of crowdsourced workers. All together, 6776 papers were accepted for publication in these 16 conferences\footnote{NAACL was not held in 2017 and 2020; it is skipped every three years.}. In total, we identified 703 papers that use crowdworkers as part of their research.\footnote{MTurk is used in 80\% of tasks.} The results are summarized in Table \ref{tab:stats}. 

\paragraph{Renumeration} For each paper that uses crowdsourced labor, we checked whether the authors discuss
payment or labor-related issues.
 Out of the 703 papers, 122 (17\%) discuss payment, either by detailing the amount paid per task, 
 the worker's hourly wages, or declaring that wages were paid ethically. While in some cases researchers emphasize fair payment (e.g., \citet{nangia-etal-2020-crows} ensured ``a pay rate of at least \$15/hour''), many other papers are more concerned about
the cost of dataset acquisition, and thus mention the cost per task or the total dataset cost,  but not the hourly compensation.

\paragraph{IRB Review} Finally, we also checked whether authors mention a review by an IRB (or equivalent body) 
for their research. We found very few papers that mention an IRB approval or exemption --- a total of 14 papers --- which make up only 2\% of the works that use crowdsourcing.

\paragraph{Growth} We see that research papers using crowdsourced tasks
have made up
a relatively constant 11-12\%
of  all research
papers  in the last three years.
As research production  grows exponentially, we expect a corresponding increase in the number of crowdsourced tasks.\footnote{ACL, EMNLP, and NAACL are only three of many other publication venues in the NLP community.}

\subsection{Categorizing Crowdsourced Tasks}
To understand the many nuanced ways in which  researchers presently use crowdworkers in NLP tasks, we examined the tasks performed in 
each of the papers that use crowdsourcing.
We found that NLP crowdsourced tasks generally fall into one of
three categories, which we designate as \textit{labeling}, \textit{evaluation}, and \textit{production}.
We found that the categories account for 34\%, 43\%, and 23\% of 
crowdsourcing tasks respectively. 
The results are summarized
in Table \ref{tab:tasks}, which also lists common action verbs that researchers  use to describe the work performed by  crowdworkers.
\input{table2}

\textbf{Labeling} entails the processing of existing data by the crowdworker and then the selection or composition of a label or labels for that data\footnote{We refrain from using the terms ``annotation'' and ``annotators'' as 
these terms are overloaded and often used for non-annotation work.}. Labeling tasks augment
 the data with human-supplied labels. The augmented data are often used for training machine learning models. We further divide labeling tasks into two: objective and subjective labeling.
In objective labeling, the desired label is factual, and does not depend on the worker.
Classical examples are the tagging of sentences with named entities and  part-of-speech (POS) labeling, 
and text transcription.
In contrast, subjective labeling comprises of tasks where labels
may depend on the worker's personality, cultural background, opinion, or affective state.
Examples include emotion labeling, detecting sarcasm in a tweet, or deciding whether a text constitutes hate speech or not.

In \textbf{evaluation} tasks, the worker is presented with data, for example a sentence, tweet, paragraph, or dialogue, and then requested to evaluate and score the data --- mostly text --- according to
predefined criteria, such as fluency, coherence, originality, or structure.
These tasks are often used by researchers to evaluate natural language generation (NLG) models.
Similar to subjective labeling, scores given by the workers may reflect the interpretation,  values, or beliefs of the worker. 

Finally, in \textbf{production} tasks, workers are asked to produce their own data, 
rather than label or evaluate existing data. In the NLP context, this often amounts to text elicitation or text generation. Examples include captioning a photo or video clip, writing a story given a sequence of images, or composing questions and answers. In this category we also include text translation. The produced data is often used for model training or evaluation.

While the majority of the studies use only one type of task --- labeling, evaluation, or production --- we found that
in 10\% of the papers that use crowdsourcing, researchers used two or more types of tasks in the same study. The combination of production and evaluation is particularly common; researchers often ask workers to generate data, which in turn is used to
train a model; they then use workers to evaluate the model's performance.

\subsection{Surveys and Gamification}
Although not common, some papers also collect personal information from workers. For example,
\citet{yang-etal-2015-using} and \citet{ding-pan-2016-personalized} conduct personality surveys among its crowdworkers. \citet{perez-rosas-mihalcea-2015-experiments} collect demographic data from
the workers which included ``their gender, age, country of origin, and education level''. 
Finally, we also found a few papers that add elements of gaming to their crowdsourced tasks, e.g. \citet{niculae-danescu-niculescu-mizil-2016-conversational} and \citet{urbanek-etal-2019-learning}.

\section{The Rules and Institutions of Research Ethics}
\label{sec:ethics}

Given the increasing use of crowdsourced NLP tasks, how can researchers ensure ethical concerns are reasonably addressed? Should a researcher make a judgement call and decide which tasks pose a risk of harm to the worker, and which are benign? To answer such questions, we will first explore the existing ethical framework used by researchers in the biomedical, social, and behavioral sciences.

\subsection{The Genesis of Modern Research Ethics}
\label{sec:genesis}
The roots of contemporary research ethics originate in the 19th century, when researchers made unparalleled discoveries,
but also engaged in hazardous, and frequently deadly, experimentation without great concern for the human subjects involved as long as these trials advanced the boundaries of scientific knowledge \cite{RN43}. The dominant ethics paradigm at the time was largely devoid of now-common principles surrounding therapeutic benefits, scientific validity, full knowledge, or subject consent \cite{RN45}. Examples include researchers infecting intellectually disabled orphans with gonorrhea, or puncturing a healthy and unaware woman with the nodules of a leper patient to observe the clinical course of these diseases \cite{RN47}.

Such incidents were common before the 1940s, and academia and public discourse were generally ignorant of them and of research ethics in general \cite{RN49}. The revelation of the Nazi concentration camp experiments at the end of World War II was a watershed moment \cite{RN48} and led to an early precursor of contemporary research ethics, namely the \textit{Nuremberg Code} of 1947 \cite{RN42}. 
Not long after, the fallout of the Tuskegee Syphilis Study in the US prompted the formalization of research ethics at American universities \cite{RN147}. In the study, which took place between 1932 and 1972, 
a total of 399 
socio-economically disadvantaged African-American males with latent syphilis infections were recruited through the false promise of “free healthcare”. Yet, these subjects were actually left without therapy even as effective treatment became available, with the objective to observe the clinical course of the disease \cite{RN145}.

\subsection{The Belmont Principles and IRBs}
From thereon, beginning with biomedical research, the notion of research ethics has been institutionalized in the US 
at the university level through institutional review boards (IRBs), as well as national legislation \cite{RN52}. 
Gradually, research ethics have become a concern also in the social and behavioral sciences, as demonstrated by the 1978 \textit{Belmont Report} created by the National Commission for the Protection of Human Subjects of Biomedical and Behavioral Research \cite{RN221}.  This report became the basis of the 1991 Federal Policy for the Protection of Human Subjects in the United States, more commonly known as the \textit{Common Rule} \cite{RN53} and superseded by the \citet{cfr}. Chiefly, the \cfr{} aims to ensure that the following three basic principles listed in the Belmont Report \cite{RN400} are met:
\begin{enumerate}
    \item     \textbf{Respect for persons}, which includes ``the requirement to acknowledge autonomy and the requirement to protect those with diminished autonomy'';

    \item \textbf{Beneficence}, which mandates considering whether the  benefits resulting from the research can outweigh the risks; and

    \item \textbf{Justice}, requiring that the burden --- and benefits --- of the research are equally shared among potential subjects.
\end{enumerate}
The \cfr{} is codified in Title 45, \textit{Code of Federal Regulations}, Part 46 and applies to all government-funded research in the United States \cite{RN52}. Virtually all universities in the United States apply this regulation to human subjects research projects irrespective of funding source \cite{RN107}. 
Specifically, the \cfr{} requires that most \textit{research involving human subjects} receives approval from an IRB. The IRB is a special university-level committee that reviews research proposals to verify that they comply with ethical standards. While it is difficult to assess the effectiveness of IRBs, and the process of ethical review is sometimes criticized as overly bureaucratic which may hamper low-risk social science \cite{resnik2018ethics,RN13}, glaring ethical lapses have been rare after 1974 \cite{RN107}.

\subsection{Are IRBs Universal?}
The three ethical principles outlined by the Belmont Report — Respect for persons, Beneficence, and Justice — are also the stated principles guiding the actions of more recently formed ethics boards around the world, as
well as the underlying principles of relevant policies of intergovernmental organizations, including the
Universal  Declaration  on  Bioethics and Human Rights \cite{unesco}
and the International Ethical Guidelines for Biomedical Research Involving Human Subjects \cite{cioms}. Consequently, many countries worldwide have modeled national regulations after the \cfr{} or its predecessor, the \textit{Common Rule} \cite{RN171}. Both regulations have also influenced editorial policies of academic journals (e.g., the \citealp{COPE}). The \cfr{} can thus be considered a cross-disciplinary de-facto standard for research ethics in Western/US-influenced academic settings \cite{Gontcharov2018}, including  India, Japan, Korea, and Taiwan.

Though we find that a global agreement on human-subjects ethics is emerging, countries still vary in
the extent to which relevant policies are accepted, framed, implemented, or enforced.
\section{Do NLP Tasks Constitute \textit{Research Involving Human Subjects}?}
\label{sec:do_nlp}
These formal rules and institutions have been established to protect the rights and interests of humans subjects involved as participants in scientific research. Thus, we must detemine whether the rules and institutions of research ethics are even applicable to crowdsourced studies. At the core of this determination are two fundamental questions: 
\begin{enumerate}
    \item Are crowdsourcing tasks \textit{research}? 
    \item Are crowdworkers \textit{human subjects}?
\end{enumerate}
In the following, we address these two questions.
\subsection{Are Crowdsourcing Tasks \textit{Research}?}
The \cfr{} defines \textit{research} as follows:
\begin{displayquote}[45 \textit{CFR} 46.102(l), \citealp{cfr}]
\textbf{Research} means a systematic investigation, including research development, testing, and evaluation, designed to develop or contribute to generalizable knowledge. Activities that meet this definition constitute research for purposes of this policy, whether or not they are conducted or supported under a program that is considered research for other purposes.
 \end{displayquote}

From this definition it is evident that rather than a concrete research behavior on part of the NLP researcher, it is the \textit{purpose} of the research behavior that classifies said behavior as \textit{research} under the \cfr{}. In other words, all categories of crowdsourced tasks summarized in Section \ref{sec:growing}, i.e., labeling,  evaluation, and  production, may be considered part of \textit{research} so long as the intended outcome is to create generalizable knowledge. Typically, this encompasses academic settings where research behavior takes place (course assignments by students being a prominent exception), but does not include research conducted in industry settings \cite{meyermichelle1,jackman2015evolving}.
\subsection{Are Crowdworkers \textit{Human Subjects}?}
\label{subsec:humansubjects}
The \cfr{} defines \textit{human subjects} as follows:
\begin{displayquote}[45 \textit{CFR} 46.102(e)(1), \citealp{cfr}]
\textbf{Human subject} means a living individual \textit{about whom} an investigator (whether
professional or student) conducting research: 
\begin{enumerate}[(i)]
\item{Obtains information or biospecimens through intervention or interaction with the
individual, and, uses, studies, or analyzes the information or biospecimens; or}
\item{Obtains, uses, studies, analyzes, or generates identifiable private information or
identifiable biospecimens. }
\end{enumerate}
\end{displayquote}
Clearly, if the researcher obtains identifiable private information (IPI) as part of the crowdsourced task ---
e.g. name, date of birth, email address, national identity number, or any other information that identifies the worker --- then (ii) holds and the worker is considered a human subject. 

Even if the researcher does not obtain any IPI, the worker may still be considered a human subject under (i) in certain cases.
This is because NLP researchers \textit{interact} with crowdworkers through MTurk or analogous platforms when they publish the task, and obtain information through this interaction. 
It is also evident that academic NLP researchers make use of this information as they conduct a given study, and hence it is ``used, studied, or analyzed''. If the information is \textit{about the crowdworker} then they are considered 
human subjects and (i) is met. However, \cfr{} does not expand on what constitutes \textit{information about the individual}. 
According to \citet{uw-worksheet}, for example, \textit{about whom} means that the ``data or information relates to the person. Asking what [crowdworkers] think about something, how they do something, or similar questions usually pertain to the individuals. This is in contrast to questions about factual information not related to the person.''

Whether the information obtained in an NLP task is about the worker
can initially seem like an easy-to-answer question. For example, \citet{benton2017ethical} write: 
\begin{displayquote}[][]
[R]esearch that requires the annotation of corpora for training models involves human annotators. But since the research does not study the
actions of those annotators, the research does not
involve human subjects. 
By contrast, if the goal
of the research was to study how humans annotate data, such as to learn about how humans interpret language, then the research may constitute human subjects research.
\end{displayquote} 
However, we believe that this is not so clear-cut.
First, one might argue that although in  a POS labeling task we do not obtain information about the worker,
other labeling tasks might be harder to classify.
For example, when researchers ask a worker to compose a story given a sequence of photos, do
they obtain information about the worker? And if so, what kind of information?
Similar questions can be asked about tasks related to emotion classification (which might reveal
a worker's personality or mood), composing questions and answers (which point to areas of interest
and cultural background), or
identifying hate speech (which can indicate political orientation).

Second, platforms might \textit{automatically}
provide information that can be considered
by some to be about the individual. 
Even in the most benign and ``objective'' tasks such as POS tagging, MTurk supplies
researchers with information on the amount of time taken to complete each task. This information is sometimes collected and used by NLP researchers (e.g., \citealp{sen2020human}).

In summary, we have shown that in an academic context, NLP crowdsourcing tasks are \textit{research}, but that the categorization of crowdworkers as \textit{human subjects} can, in some cases, be a gray
area that is open to interpretation. 
The \cfr{} was designed to address ethical issues in medical research, and later in 
behavioral sciences; lawmakers and experts involved did not anticipate its use in new domains
such as crowdsourcing. Therefore,  its application to online
data collection, and crowdsourcing in particular, can be ambiguous and unsatisfactory.\footnote{Some universities employ the \textit{Precautionary Principle} and require all crowdsourced-enabled research to go through an IRB application.}
Thus, while in some cases the protections and procedures mandated under the \cfr{} apply,
in others they might not. As a consequence, some NLP crowdsourcing tasks may not require an IRB application,
and this may happen even if crowdworkers are at risk. 

\section{Dispelling IRB Misconceptions}
\label{sec:points}
As we now see, 
if workers constitute human subjects, an IRB application is required. 
To clarify any other misconceptions that might be present within the community, we list key points related
to the IRB process and dispel misconceptions around them.
While not exhaustive, the list can serve both researchers and reviewers.
\subsection{Researchers Cannot Exempt Themselves from an IRB Application} 
The \cfr{} includes provisions for IRB exemptions, and we expect the vast majority of crowdsourced NLP tasks to fall into that category. However, it is crucial to understand that granting a research project an IRB exemption is not the prerogative of the researcher; it is only  IRB that hands out exemptions following an initial review: 
\begin{displayquote}[\citealp{aapor}][]
[T]he determination of exempt status (and the type of review that applies) rests with the IRB or with an administration official named by your institution. \textit{The determination does not rest with the investigator.} Therefore, all projects must be submitted to the IRB for initial review.
\end{displayquote}

\subsection{Worker IDs Constitute IPI}
\label{sub:workerid}
Researchers often obtain and store worker IDs ---   unique and
persistent identifiers assigned to each worker by the crowdsourcing platform --- even when workers are ``anonymous''. MTurk, for example, assigns each worker a fixed 14-digit string, which is provided to the researcher with completed tasks. 
A worker ID is part of the worker's account, and is therefore linked to their personal details, including full name, email address, and bank account number.
As a consequence, the worker ID constitutes IPI (identifiable private information). If the worker ID is obtained by the researcher, the research mandates an initial IRB review.

To avoid obtaining this IPI, researchers can create and store pseudonymized worker IDs, provided that these IDs cannot be mapped back to the original worker IDs.
\subsection{Obtaining Anonymous Data Does Not Automatically Absolve from IRB Review}
Even if IPI is not obtained, and only \textit{anonymous} and non-identifiable data is collected, we have shown
in Section \ref{subsec:humansubjects} that crowdsourced NLP tasks involve interaction between researcher and participant through which the data about the worker may be collected, and thus often require an initial review by an IRB. 

\subsection{Payment to Crowdworkers Does Not Exempt Researchers from IRB Review}
\label{subsec:payment}
Remuneration of human subjects does not change their status to independent contractors beyond the scope of research ethics. In fact, compensation of human subjects for the time and inconvenience involved in participating is a standard practice ``especially for research that poses little or no direct benefit for the subject'' and at the same time ``should not constitute undue inducement to participate'', as the \citet{uofvirginia} points out.

\subsection{Non-Published Research Also Requires IRB Review} Some researchers believe that research that will not be published is not subject to an IRB review. For example,  \citet{cmu-course} teaches students that ``Paid labeling does not require IRB approval... [b]ut sometimes you want to discuss results in papers, so consider IRB approval''.\footnote{Moreover, whether the labeling is paid or unpaid is irrelevant; see Section \ref{subsec:payment}.}

The definition of \textit{research} in the \cfr{} is not contingent on subsequent publication of the results. Given the uncertainties of the peer-review process, whether research eventually finds its way into a publication can often be ascertained only \textit{ex post}. An important exception not requiring IRB approval is student work as part of course assignments. However, subsequent use of research data collected originally as part of a student assignment is not mentioned in the \cfr{}.
Consequently, universities handle this situation differently. For example,  University of Michigan allows retroactive IRB approval:
\begin{displayquote}[\citealp{umich}]
Class assignments may become subject to this policy... if the faculty member or the students change their plans... application to the IRB for permission to use the data is required. 
\end{displayquote}
while Winthrop University does not:
\begin{displayquote}[\citealp{uwinthrop}]
    IRB approval cannot be granted retroactively, and data may need to be recollected for the project.
\end{displayquote}

\section{Risks and Harms for Crowdworkers}
\label{sec:risks}
Previous work on the ethics of crowdsourcing focused on labor conditions, such as fair pay, and on privacy issues \cite{fort2011amazon,gray2019ghost}.
However, even when payment is adequate and the privacy of the workers is preserved, there are additional ethical considerations that are often not taken into account, and might put the worker at risk. 
We propose using the three ethical principles outlined by the Belmont Report --- Respect for Persons, 
Beneficence, and Justice --- to guide the action of researchers. 
We outline some of the specific
risks and harms that might befall NLP task crowdworkers in light of these principles.
While the list is not comprehensive, it can serve as a starting point to be used by researchers when planning their crowdsourced task, as well
as by reviewers examining the ethical implications of a manuscript or research proposal.

\subsection{Inducing Psychological Harms} 

NLP researchers are increasingly cognizant that texts can potentially harm readers, 
as evident by \textit{trigger warnings} they add to their own papers \cite[e.g.,][]{sap-etal-2020-social,nangia-etal-2020-crows,sharma-etal-2020-computational, han-tsvetkov-2020-fortifying}. 
Moreover, researchers have long known that annotation work may be psychologically harmful.
The Linguistic Data Consortium (LDC), for example, arranged stress-relieving activities for annotators of broadcast news data, following reports of ``negative psychological impact'' such as intense irritation, overwhelmed feelings, and task-related nightmares \citep{strassel-etal-2000-quality}. Although literature on
the emotional toll on crowdworkers is still scant \cite{huws2015}, there is growing literature on the psychological cost of work done by commercial content moderators \cite{steiger2021psychological}. Crowdworkers deserve similar consideration: while NLP tasks can be as benign as the POS tagging of a children's poem, 
they may also involve exposure to disturbing textual or visual content.

In 45 \textit{CFR} 46.110 the \cfr{} allows expedited IRB review for research posing \textit{no more than minimal risk} to the human subjects involved. According to 45 \textit{CFR} 46.102(j) \textit{minimal risk} ``means that the probability and magnitude of harm or discomfort anticipated in the research are not greater in and of themselves than those ordinarily encountered in daily life or during the performance of routine [...] psychological examinations or tests.'' Exposing crowdworkers to 
sensitive content may exceed the threshold of \textit{minimal risk} if the data that requires labeling or evaluation might be psychologically harmful. 

The amount of harm (if any) depends on the sensitivity of the worker
to the specific type of content. The risk is generally 
higher in data labeling or text evaluation tasks, 
where workers might be repeatedly asked to categorize
offensive tweets, transcribe violent texts, evaluate hateful text that may expose them to emotional stimuli, or, depending on the content and worker, traumatize them. 
In some cases, sexual material can be highly offending or shocking and cause an emotional disturbance. Although less likely, hazards can also occur when workers are asked to produce text, since workers are elicited to produce texts based on given input. For example, when users are asked to compose a story based on images, certain images might trigger a harmful response in the worker.

\subsection{Exposing Sensitive Information of Workers} A crowdworker might inadvertently or subconsciously expose sensitive information about
themselves to the researcher. This is more pronounced in text production, where the responses produced by such tasks reveal as much about the individual workers as they do about the produced text. However, workers also reveal information about themselves  when evaluating or labeling text, especially when subjective labeling is in place.
Moreover, even seemingly trivial data --- for example,
the elapsed time taken by the worker to label, evaluate, or produce text
--- may contain valuable information about the worker (and this information
is automatically captured by MTurk). Table \ref{tab:risk} shows the risk level for the different task categories.

Moreover, researchers can obtain sensitive information about workers because the crowdsourcing platforms allow screening of workers using built-in qualification attributes including age, financial situation, physical fitness, gender, employment status, purchasing habits, political affiliation, handedness, marital status, and education level \cite{premium}. 

Researchers may also obtain other types of sensitive information by creating their own, arbitrary qualification tests as a quality control measure \cite{qualitycontrol}.

In summary, the information obtained by researchers may reveal as much about the individual crowdworkers as they do about the data being labeled, evaluated,
or produced.
\input{table3}

\subsection{Unwittingly Including Vulnerable Populations}
Research on crowdsourcing platforms such
as MTurk is inherently prone to the inclusion of \textit{vulnerable} populations. In 45 \textit{CFR} 46.111(b) the \citet{cfr} non-exhaustively lists ``children, prisoners, individuals with impaired decision-making capacity, or economically or educationally disadvantaged persons'' as vulnerable groups. 
The \cfr{} requires that additional safeguards are implemented to protect these vulnerable populations and makes IRB approval contingent on these safeguards.

A great proportion of MTurk crowdworkers are located in developing countries, such as India or Bangladesh, which makes MTurk an attractive proposition to those offering a task \cite{gray2019ghost}, but increases the risk of including economically-disadvantaged persons. Furthermore, it is difficult to ensure that MTurk crowdworkers are above the age of majority, or fall into any other of the defined vulnerable populations \cite{Mason2011}.

Moreover, given the power imbalances between researchers in industrialized countries and crowdworkers in developing countries, ethical consideration should occur regardless of whether the jurisdiction in which the crowdworkers are located even has a legal framework of research ethics in place, or whether such a local framework meets the standard of the Belmont Report or \cfr{}.

\subsection{Breaching Anonymity and Privacy}
There is a perception among researchers that crowdworkers are anonymous and thus the issue of privacy is not a concern. This is not the case. \citet{lease2013mechanical}, for example, discuss a vulnerability that can expose the identity of an Amazon Mechanical Turk worker using their worker ID ---
a string of 14 letters and digits --- because the same worker ID is used also as the identifier
of the crowdworker's account on other Amazon assets and properties. As a result, a Google search for the worker ID can lead to personal information such as product reviews written by the crowdworker on Amazon.com, which in turn can disclose the worker's identity. Researchers might be
unaware of these issues when they make worker IDs publicly available in papers or in datasets. For example, \citet{gao-etal-2015-cost} 
rank their crowdworkers using MTurk worker IDs in one of the figures. 

Moreover, breaches of privacy can also occur unintentionally. For example,
workers on MTurk are provided with an option to contact the researcher. 
In this case, their email address will be sent to the researcher, who is inadvertently exposed to further
identifiable private information (IPI). We maintain that the anonymity of crowdworkers cannot be automatically assumed or guaranteed, as this is not a premise of the crowdsourcing platform.

\subsection{Triggering Addictive Behaviour} \citet{graber2013internet} identify another source for harmful effects, which ties in with the risk of psychological harm, and is specific to \textit{gamified} crowdsourced tasks: a possibility of addiction caused  by dopamine release following a reward given during the gamified task. Gamification techniques can be added to data labeling, evaluation, and production. Indeed, some NLP work is using gamification, mostly for data collection \cite[e.g.,][]{kumaran2014online, ogawa-etal-2020-gamification,ohman-etal-2018-creating}.

\begin{figure}[h]
\centering
\includegraphics[width=\columnwidth]{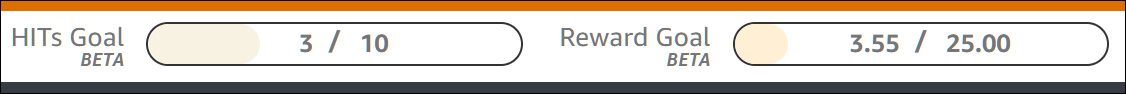}
\caption{The ``Daily Goals Dashboard'' on MTurk}
\label{fig:mturk}
\end{figure}

Moreover, the crowdsourcing platform may add
elements of gamification over which the researcher has no control.
For example, 
MTurk recently introduced a ``Daily Goals Dashboard'', where the
worker can set game-like ``HITs Goal'' and ``Reward Goal'', as shown in Figure \ref{fig:mturk}. 

\section{Ways Forward}
\label{sec:conclusion}
The use of crowdworkers is growing within the NLP community, but the ethical framework set in place
to guarantee their ethical treatment (whether \textit{de jure} or \textit{de facto}) did not anticipate the emergence of crowdsourcing platforms. 
In most crowdsourced NLP tasks, researchers do not intend to gather information about the worker. However,
the crowdsourcing platform often autonomously collects such information. 
As a result, it is often difficult to determine whether crowdworkers constitute human subjects --- which hinges on whether the researcher
collects information about the worker. 
However, a determination that all crowdworkers are human subjects --- and thus mandate an IRB approval for 
government-supported institutions --- might create a ``chilling effect''
and disadvantage university researchers compared to industry-affiliated researchers.
The effect is exacerbated in institutions where the ethics committee  is heavily bureaucratic  and does not offer a
streamlined, expedited exemption process for low-to-no risk studies.

Whether their crowdsourced task requires IRB application or not, we recommend that the ethics-aware
researcher should carefully examine their study in light of the three principles set up by the Belmont Report: Respect for persons, Beneficence, and Justice.  And while this mandates fair pay, it is important to note that this
is just one of the implications. There are other ethical considerations that are often overlooked --- in particular
risk assessment of causing psychological harm and exposure of sensitive information. Thus, we recommend increasing awareness of the potential ethical implications of crowdsourced NLP tasks. As NLP researchers are now encouraged to add an ``ethical considerations'' section to their papers \cite{naacl-ethics-statement}, they should also be encouraged to carefully weigh potential benefits against risks related to the crowdsourced task. 

We also propose increasing awareness by disseminating relevant knowledge and information. An educational ethics resource created using a community effort could serve as a beneficial first step. Such a resource can include 
guidelines, checklists, and case studies that are specific to the ethical challenges of crowdsourced tasks in the 
context of NLP research. We believe that the creation of such a resource can serve as a springboard for a 
necessary nuanced conversation regarding the ethical use of crowdworkers in the NLP community.
\section*{Acknowledgements}
We thank the anonymous reviewers for their valuable comments and suggestions which helped improve the paper.
This research was partially supported by the Ministry of Science and Technology in Taiwan under grants MOST 108-2221-E-001-012-MY3 and MOST 109-2221-E-001-015- [\textit{sic}].
\bibliography{custom}
\bibliographystyle{acl_natbib}
\end{document}

%% file: table1.tex
\begin{table*}[t!]
\centering
\resizebox{0.80\linewidth}{!}{%
\begin{tabular}{@{}c|ccc|c|ccc@{}}
\toprule
\multicolumn{1}{c}{\textbf{~}} & \multicolumn{4}{c}{\textbf{Accepted Papers}} &  
\multirow{2}{*}{\textbf{Papers Using}} &   \multirow{2}{*}{\textbf{Payment}} &  \multirow{2}{*}{\textbf{IRB}} \\ \cmidrule{2-5}
\multicolumn{1}{c}{\textbf{Year~~~}} & \multicolumn{1}{c}{\textbf{ACL}} & \multicolumn{1}{c}{\textbf{EMNLP}} &   \multicolumn{1}{c}{\textbf{NAACL}} & \multicolumn{1}{c}{\textbf{All}} & \multicolumn{1}{c}{\textbf{Crowdsourcing}} 
 & \multicolumn{1}{c}{\textbf{Mentioned}}  & \multicolumn{1}{c}{\textbf{Mentioned}}\\
\midrule
2015 & 318 & 312 & 186 & ~816  & ~~59 ~~(7\%)  &~~4 ~(7\%) & 0 \\
2016 & 328 & 264 &  182 & ~774 & ~~82 ~(11\%) & 15 (18\%) & 0 \\
2017 & 302 & 323 &  --- & ~625 & ~~57 ~~(9\%)  & ~12 (21\%) & 3 \\
2018 & 381 & 549 & 332 & 1262 & 136 (11\%)  & ~17 (13\%) & 1 \\
2019 & 660 & 683 & 423 & 1766 & 189 (11\%)  & 32 (17\%) & 5 \\
2020 & 779 & 754 & --- & 1533 & 180 (12\%)  & 42 (23\%) & 5 \\ \midrule
Total & 2768 & 2885 & 1123 & 6776 & 703 (10\%) & 122 (17\%) & 14 \\ \bottomrule
\end{tabular}%
}

\caption{Papers using crowdsourced tasks in top NLP conferences, 2015--2020. The columns show, from left to right: conference year; number of accepted papers at ACL, EMNLP, and NAACL; total number of accepted papers;  number of accepted papers using crowdsourced tasks (percentage of papers using crowdsourced tasks); 
 number of papers using crowdsourced tasks that mention  payment (percentage of papers using crowdsourced
tasks that mention  payment); number of papers using crowdsourced tasks that mention IRB review or exemption.
}
\label{tab:stats}
\end{table*}

%% file: table2.tex
\begin{table}[h]
\centering
\resizebox{1.0\columnwidth}{!}{%
\begin{tabular}{@{}l|c|l@{}}
\toprule
\textbf{Task }       & \multicolumn{1}{c|}{\textbf{}}                            &  \textbf{Action}             \\ 
\textbf{Category}       & \multicolumn{1}{c|}{\textbf{Pct.}}                            &  \textbf{Verbs}             \\ \midrule
Labeling                                  & 34\%                       
& annotate, label, identify, indicate, check \\ 
Evaluation                                      & 43\%                       & rate, decide, score, judge, rank, choose                             \\ 
Production                                    & 23\%                       & write, modify, produce, explain, paraphrase                                \\
\bottomrule
\end{tabular}%
}
\caption{Tasks in the papers that use crowdsourcing, broken down by category. We also list typical action verbs used by the authors to describe the task.}
\label{tab:tasks}
\end{table}

%% file: table3.tex
\begin{table}[tb]
\centering
\resizebox{\columnwidth}{!}{%
\begin{tabular}{@{}c|c@{}}
\toprule
\textbf{Task} & \textbf{Risk of Exposure of Workers'} \\ 
\textbf{Category} & \textbf{Sensitive Information}  \\ \midrule
{Objective Labeling}  & Low            \\
{Subjective Labeling} & Medium         \\
{Evaluation}            & Medium        \\
{Production}            & High   \\ \bottomrule
\end{tabular}%
}
\caption{Potential risk level by task category.}
\label{tab:risk}
\end{table}